\documentclass[sn-nature,Numbered]{sn-jnl}% Math and Physical Sciences Reference Style
\usepackage{graphicx}%
\usepackage{multirow}%
\usepackage{amsmath,amssymb,amsfonts}%
\usepackage{amsthm}%
\usepackage{mathrsfs}%
\usepackage[title]{appendix}%
\usepackage{xcolor}%
\usepackage{textcomp}%
\usepackage{manyfoot}%
\usepackage{booktabs}%
\usepackage{algorithm}%
\usepackage{algorithmicx}%
\usepackage{algpseudocode}%
\usepackage{listings}%
\usepackage{hhline}
\begin{document}

\title[Article Title]{A Multi-Modal Deep Learning Based Approach for House Price Prediction}


\author*[1]{\fnm{Md Hasebul} \sur{Hasan}}\email{hasebulhasan97@gmail.com}
\equalcont{These authors contributed equally to this work.}

\author*[1]{\fnm{Md Abid} \sur{Jahan}}\email{abidjahanapon@gmail.com}
\equalcont{These authors contributed equally to this work.}

\author*[1]{\fnm{Mohammed Eunus} \sur{Ali}}\email{eunus@cse.buet.ac.bd.com}

\author[2]{\fnm{Yuan-Fang} \sur{Li}}\email{yuanfang.li@monash.edu}

\author[3]{\fnm{Timos} \sur{Sellis}}\email{timos@athenarc.gr}

\affil*[1]{\orgdiv{Department of Computer Science and Engineering}, \orgname{Bangladesh University of Engineering and Technology}, \orgaddress{\city{Dhaka}, \country{Bangladesh}}}

\affil[2]{\orgname{Monash University}, \orgaddress{\city{Clayton}, \country{Australia}}}

%\affil[3]{\orgname{Swinburne University of Technology}, \orgaddress{\city{Hawthorn}, \country{Australia}}}

\affil[3]{\orgname{Archimedes Research Unit, Athena Research Center}, \orgaddress{\city{Marousi}, \country{Greece}}}


\abstract{\label{abstract}
Accurate prediction of house price, a vital aspect of the residential real estate sector, is of substantial
interest for a wide range of stakeholders. However, predicting house prices is a complex task
due to the significant variability influenced by factors such as house features, location, neighborhood, and many others. Despite numerous attempts utilizing a wide array of algorithms, including recent deep learning techniques, to predict house prices accurately, existing approaches have fallen short of considering a wide range of factors such as textual and visual features. This paper addresses this gap by comprehensively incorporating attributes, such as features, textual descriptions, geo-spatial neighborhood, and house images, typically showcased in real estate listings in a house price prediction system. Specifically, we propose a multi-modal deep learning approach that leverages different types of data to learn more accurate representation of the house. In particular, we learn a joint embedding of raw house attributes, geo-spatial neighborhood, and most importantly from textual description and images representing the house; and finally use a downstream regression model to predict the house price from this jointly learned embedding vector. Our experimental results with a real-world dataset show that the text embedding of the house advertisement description and image embedding of the house pictures in addition to raw attributes and geo-spatial embedding, can significantly improve the house price prediction accuracy. The relevant source code and dataset are publicly accessible at the following URL: https://github.com/4P0N/mhpp}




\maketitle

\section{Introduction}
\label{introduction}

The global economy is significantly influenced by the real estate market. The global real estate market is projected to attain an impressive value of US\$637.80 trillion by the year 2024. (source: Statista~\footnote{https://www.statista.com/outlook/fmo/real-estate/worldwide}). Across many countries, owning a home is considered a major life milestone and represents the most valuable asset in an individual's life. Consequently, the evaluation of house prices, a vital aspect of the residential real estate sector, holds substantial interest for a wide range of stakeholders, including prospective buyers, sellers, real estate professionals and financiers. However, predicting house prices is a complex task due to the significant variability influenced by factors such as property characteristics, location, neighborhood attributes and many others. In this paper, we introduce an innovative approach based on multi-modal deep learning, which effectively utilizes available data from typical real estate company websites, encompassing details such as house attributes, descriptions, images and more, to achieve highly accurate house price predictions.

Considering the huge impact of this problem domain in real life, many house price prediction methods have been proposed over the years, e.g.,\cite{owusu2011review}. Earlier methods such as the Hedonic Price models \cite{rosen1974hedonic} used different house features such as the number of bedrooms, kitchens, balconies, washrooms, etc.\ to predict house prices. Later models incorporate spatial features along with house features to predict prices~\cite{bourassa2003housing,fik2003modeling,bourassa2007spatial}. However, these approaches suffer from the limitations of explicit feature engineering that require the direct involvement of domain experts. To alleviate this limitation, several machine learning and deep learning model techniques have been proposed~\cite{wang2014real, piao2019housing,zhao2019deep,chen2017house}. %\cite{zhao2019deep} used images to esitmate the price of the house.
Among these techniques, the recently proposed Geo-Spatial Network Embedding (GSNE) method~\cite{das2021boosting} shows that by incorporating key spatial and neighborhood features such as schools and train stations, it can significantly improve the house price prediction accuracy.

\begin{figure}
        \centering
   
            \includegraphics[width=0.7\textwidth]{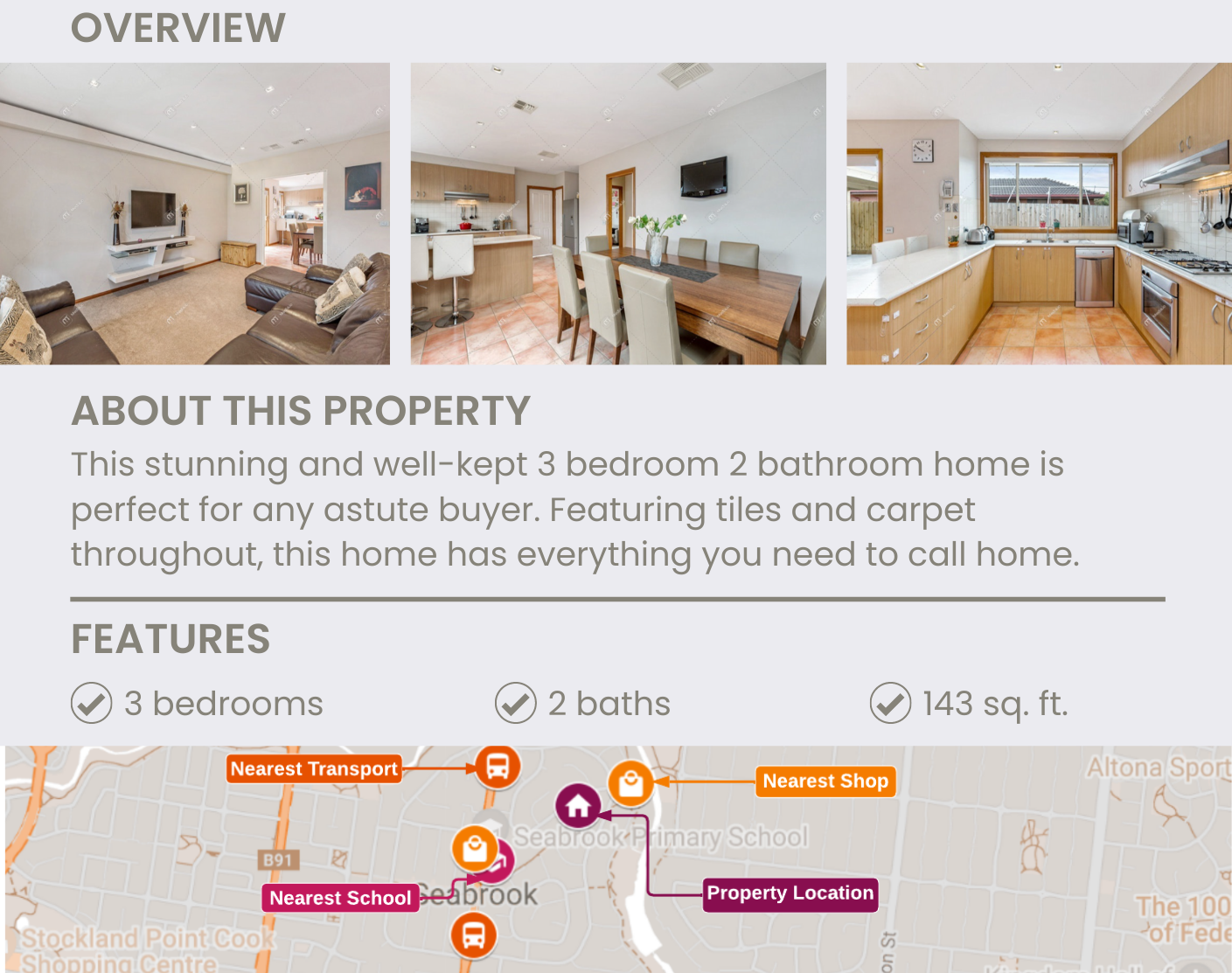}
            \label{fig:houseflyer}
            \caption{ A simple house flyer and its contents}
\end{figure}

The recent successes of multi-modal deep learning techniques in the fields of computer vision and natural language processing (NLP)~\cite{devlin2018bert,vaswani2017attention,li2021supervision,gomez2017self} have inspired us to explore the utilization of additional house-related features to further enhance the accuracy of house price predictions. We have observed that typical real estate websites provide information including house features, the surrounding neighborhood (including location and points of interest), a concise textual description and a collection of images showcasing the property. These elements collectively assist users in making informed decisions when selecting a suitable house. Figure~\ref{fig:houseflyer} illustrates a typical real estate agent's flyer for a house, featuring four key components: (1) house features (e.g., number of bedrooms, bathrooms, area), (2) spatial neighborhood features (e.g., location, proximity to schools, etc.), (3) textual descriptions and (4) house images. In this paper, we present a multi-modal deep learning-based framework for house price prediction, which leverages information from all these four data types for a highly accurate house price predictions.

More specifically, we use four separate models to learn the embeddings of raw house features, spatial neighborhood, textual description and images and concatenate the learned embedded vectors of these four streams into a single feature vector. This final feature vector is used to predict the house price using different downstream machine learning models. Though the first two types of features have been used in~\cite{das2021boosting}, in this paper, we adopted the well-known Transformer based BERT to learn the embedding of house description and exploit the multimodal language model, CLIP~\cite{li2021supervision, gomez2017self}, to jointly learn the embedding of text description and images of the house. We name our approach, the Multi-Modal House Price Predictor (MHPP). %Finally, we have experimented wth a real-world dataset collected

Finally, we have tested the effectiveness of our proposed approach with a real-world real-estate dataset of Melbourne, Australia, consisting of 52,851 house sale transaction records between 2013 and 2015. The experimental results show that the incorporation of the embeddings of the textual description and images of the house into the raw features and geo-spatial embeddings can significantly improve the house price prediction accuracy.

In summary, the key contributions of this paper are as follows.

\begin{itemize}
    \item We propose a Multi-Modal House Price Predictor (MHPP) of the house that can learn more accurate representations of the house features from joint embeddings of raw house features, geo-spatial neighborhood and most importantly from textual description and images representing the house. 
    \item We employ a pre-trained language model to use the textual description of a house and use the learned text embedding as a feature for the house price prediction.
    \item We adopt contrastive learning for multi-modal feature learning by jointly learning the house images and the house description to better capture the visual aesthetics of the house. 
    \item  We have conducted an extensive experimental study with a real-world real estate dataset, which demonstrates the effectiveness of our proposed MHPP on a wide range of downstream machine learning models.
\end{itemize}
\section{Related Work}
\label{related work}
House price prediction has garnered substantial attention within the research community, prompting a thorough exploration of diverse methodologies. This investigation has primarily unfolded across three principal categories: location-centric methodologies, machine learning-based strategies and conventional methodologies that emphasize dwelling attributes.

\textbf{Traditional Approaches}:
Early endeavors in predicting house prices were rooted in the principles of hedonic regression~\cite{rosen1974hedonic}. Subsequent research endeavors expanded upon this foundational model to forecast prices across distinct markets and scrutinize the impacts of various variables \cite{trojanek2013measuring,yayar2014hedonic, krol2015application, ottensmann2008urban}. The hedonic price model conceptualizes houses as amalgams of multiple attributes, with consumers purchasing bundles of these attributes. However, while this method simplifies the prediction process, it exhibits certain limitations. Notably, hedonic pricing coefficients for specific features display instability across diverse locations, property types and ages \cite{fletcher2000modelling}. Furthermore, issues pertaining to model specification, interactions among independent variables, non-linearity and the presence of outlier data points curtail the efficacy of hedonic price models \cite{limsombunchai2004house}. Genetic algorithms have also been explored in the context of predicting home prices, where a hybrid approach combining genetic algorithms and Support Vector Machines (SVM) has been deployed \cite{ng2008using}, alongside another study investigating the potential of genetic algorithms in this domain \cite{manganelli2015using}. Furthermore, genetic algorithms have been harnessed to examine the influence of geographic location on various outcomes, with evolutionary polynomial regression recently applied to housing price modeling \cite{morano2018multicriteria}.

\textbf{Machine Learning Approaches}:
Acknowledging the achievements of machine learning models across diverse prediction tasks, researchers have embraced a spectrum of machine learning approaches for home price prediction. Support Vector Machine (SVM) regression was employed to compute house prices in \cite{wang2008application} and \cite{li2009svr}, while Lasso and Ridge Regression were enlisted for price prediction in \cite{xin2018modelling}. Artificial Neural Network (ANN)-based models have exhibited superior performance compared to hedonic price models in out-of-sample predictions \cite{limsombunchai2004house}. A comprehensive study evaluated multiple machine learning methods, encompassing ANN, AdaBoost, Random Forest, Gradient Boosted Trees, Multi-Layer Perceptron and ensemble learning algorithms, determining Gradient Boosted Trees as the most effective in predicting home prices \cite{chaturvedi2021real}. Deep learning techniques, such as Convolutional Neural Networks (CNN), have been deployed for feature selection and price prediction \cite{piao2019housing}. A comparative analysis of Multi-Level Modeling (MLM) strategies against ANN techniques underscored the supremacy of MLM techniques \cite{feng2015comparing}. In a recent study~\cite{zhao2019deep}, tabular features were integrated with property photos using CNN to extract features from images and these features were then merged with modified tabular features. The resulting integration, when combined with the XGBoost algorithm, led to enhanced performance in price prediction. However, the model did not incorporate any textual description or house location for predicting house prices, which is a significant limitation addressed in our work.

\textbf{Location-Centric Approaches}: 
Recognizing the significance of spatial factors in prediction models for accounting for location's impact on home prices, researchers have advanced methodologies that address spatial dependence. Conventional hedonic regression models~\cite{rosen1974hedonic} operate under the assumption of relative independence among residuals. However, empirical observations have identified spatial dependence in these residuals~\cite{pace1998generalizing}. Several strategies have been proposed to incorporate spatial dependence, including methods that combine residuals from proximate properties with independent submarket equations~\cite{bourassa2003housing} and models employing postcode dummies as predictive variables ~\cite{fletcher2000modelling}. Geo-statistical techniques have demonstrated superior performance compared to Ordinary Least Squares (OLS) regression models~\cite{dubin1998predicting,basu1998analysis}. Geostatistical models incorporating disaggregated submarket variables have yielded the most accurate price predictions while accounting for spatial dependency~\cite{bourassa2010predicting}. At the neighborhood level, geostatistical models exhibited marginal improvements over OLS models~\cite{thibodeau2003marking}.In a recent state-of-the-art study (GSNE)~\cite{das2021boosting}, the intrinsic relationships between neighborhood points of interest and associated homes were investigated. 

 House pictures and text descriptions were recognized as essential elements for accurately predicting home prices. While previous efforts~\cite{8836731,poursaeed2018vision} focused on house photos or textual descriptions for house price prediction, none of them integrated all relevant elements, including home raw features, pictures, text and geospatial context, into a comprehensive approach. In this study, we consolidate all these factors to accurately anticipate home prices and leverage their combined predictive capacity.
\section{Methodology}
\label{methodology}
In this section, we provide a comprehensive description of our methodology MHPP. Our MHPP model is designed to forecast house prices by leveraging a rich set of features, including raw house feature, geospatial context, textual descriptions and images of the house.  The raw features encompass factors like room count, bathrooms, balconies, gardens and more. The geospatial context accounts for the impact of location and nearby points of interest (POIs) such as schools, train stations, colleges, shopping malls and more on house prices. 

Our prediction system begins by extracting feature information from the geospatial context, textual description and images of houses. We refer to this extracted feature information as ``embedding". Subsequently, we concatenate this embedding with the raw house features and feed this combined representation into a downstream regression model to predict the house price.

\begin{figure}
        \centering 
            \includegraphics[width=0.9\textwidth]{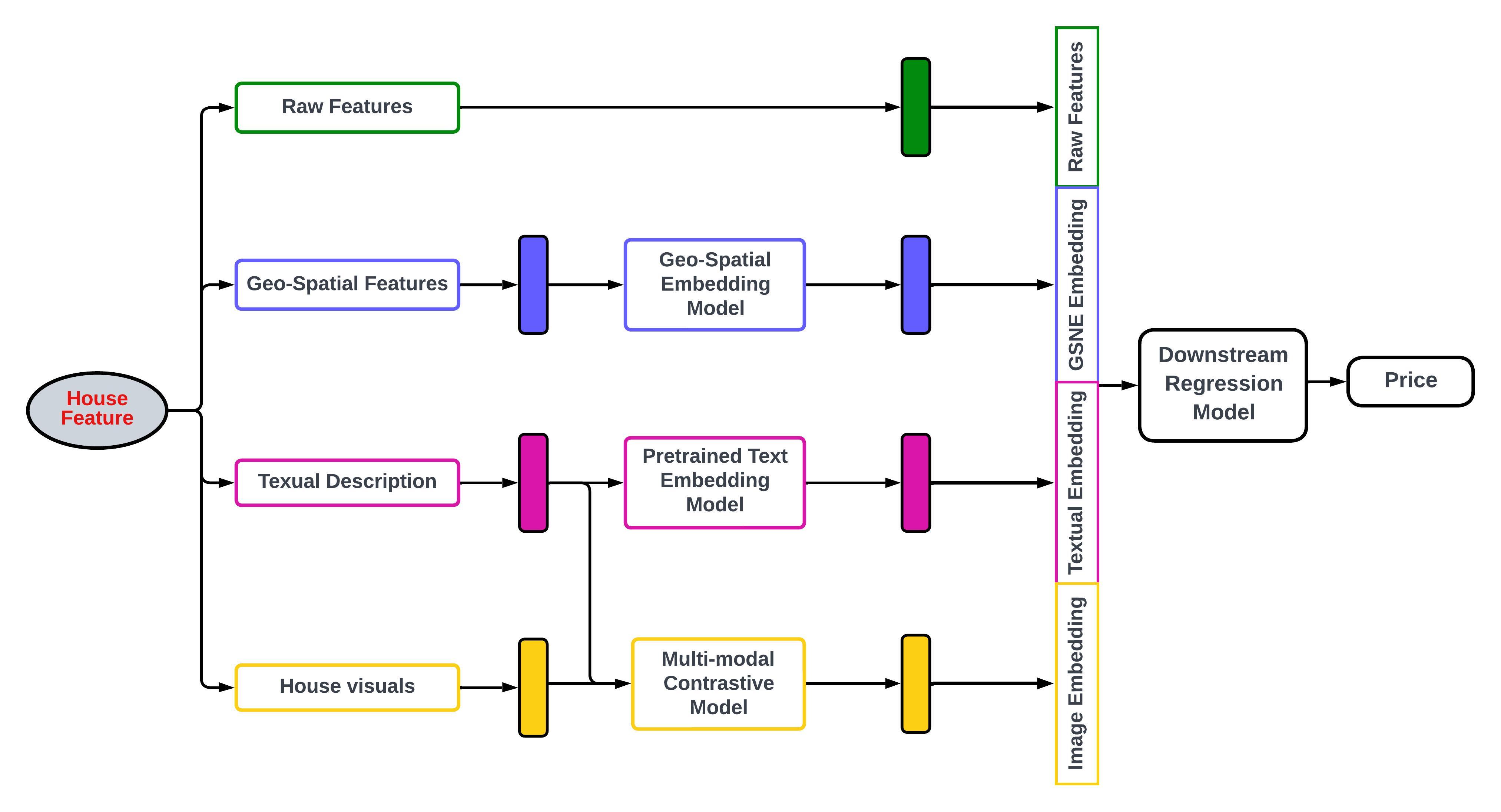}
        \caption{ MHPP joint embedding, incorporating all relevant house features.}
        \label{fig:overall}
\end{figure}

In Figure \ref{fig:overall}, the extraction of embeddings from various house features is demonstrated. Specifically, three separate models are utilized for this purpose. Geospatial features are input into a dedicated spatial embedding model known as GSNE, textual descriptions are fed into a pretrained text embedding model and house images are processed through a multi-model contrastive model to derive image embeddings. Subsequently, these three distinct embeddings are combined with the raw features to create the final feature vector that is used for predicting house prices.

\subsection{Geo-spatial Context Embedding}
\label{gsne}
The location of a house, along with proximity to key points of interest such as train stations, bus stops and schools, plays a crucial role in determining the house price. A recent study, Geo-Spatial Network Embedding(GSNE) method introduced in \cite{das2021boosting}, recognized this significance and developed an embedding technique that incorporates the geographical context of houses with the raw house features. In our research, we employed the GSNE to extract geo-spatial neighborhood features embedding for predicting house prices.

\begin{figure}
        \centering
           \includegraphics[width=0.8\textwidth]{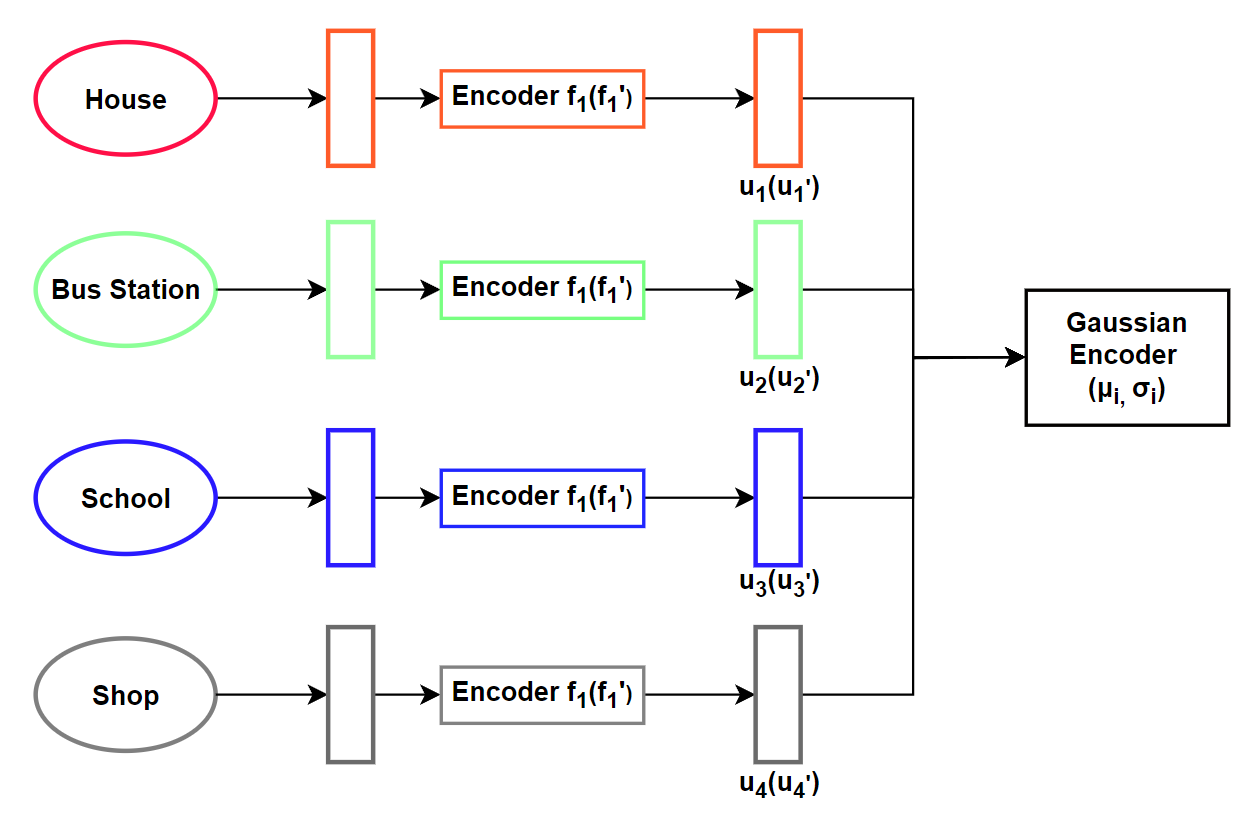}

        \caption{GSNE architecture overview.}
        \label{fig:gsne}
\end{figure}

Figure \ref{fig:gsne} provides a visual representation of the comprehensive framework of the GSNE architecture. The nodes within this geo-spatial graph represent houses and a specific subset of POIs, encompassing entities like schools, bus stations, or shops. The edges within this graph are forged based on spatial characteristics, connecting pairs of nodes if their distance falls below a user-defined threshold, denoted as $\delta_{max}$. For each edge $e_{ij} \in E$, its weight is computed as $w_{ij} = \frac{1} {\delta(i,j)}$, where $\delta(i,j)$ quantifies the Euclidean distance between nodes $i$ and $j$.

The GSNE  leverages both the first and second-order proximity. The first-order proximity ($f_k$) captures information from direct connections between node pairs across different partitions, while the second-order proximity ($f_k'$) gleans insights from nodes connected through intermediary nodes. This dual approach enables GSNE to capture both local neighborhood context and global network connectivity. Subsequently, the intermediate representations of node $i$ following the attribute encoding stage, denoted as $u_i$ for the first-order proximity and $u_i'$ for the second-order proximity, are processed through a Gaussian encoder to yield the ultimate Gaussian embedding, denoted as $\mathcal{N}(\mu_i,\sigma_i)$.

It is noteworthy that a common global Gaussian encoder, specifically ($f_{\mu},f_{\sigma}$) for the first-order proximity and ($f_{\mu}',f_{\sigma}'$) for the second-order proximity, is employed to project all nodes into the same $L$-dimensional Gaussian space. The Gaussian embedding $G_E$ is generated using the following equations:

\begin{equation}
{\bf \mu}i = f{\mu}(u_i) = ReLU(W_{\mu}u_i + b_{\mu})
\end{equation}
\begin{equation}
{\bf \sigma}i = f{\sigma}(u_i) = ReLU(W_{\sigma}u_i + b_{\sigma}) + 1
\end{equation}

In these equations, $W_{\mu}$ and $W_{\sigma}$ represent the mean and covariance encoder weights for the first-order proximity, while $b_{\mu}$ and $b_{\sigma}$ signify the mean and covariance encoder biases for the first-order proximity. These equations hold true for the second-order embedding as well.

\begin{figure}
    \centering
    %\begin{subfigure}{0.99\textwidth}
        \includegraphics[width=\textwidth]{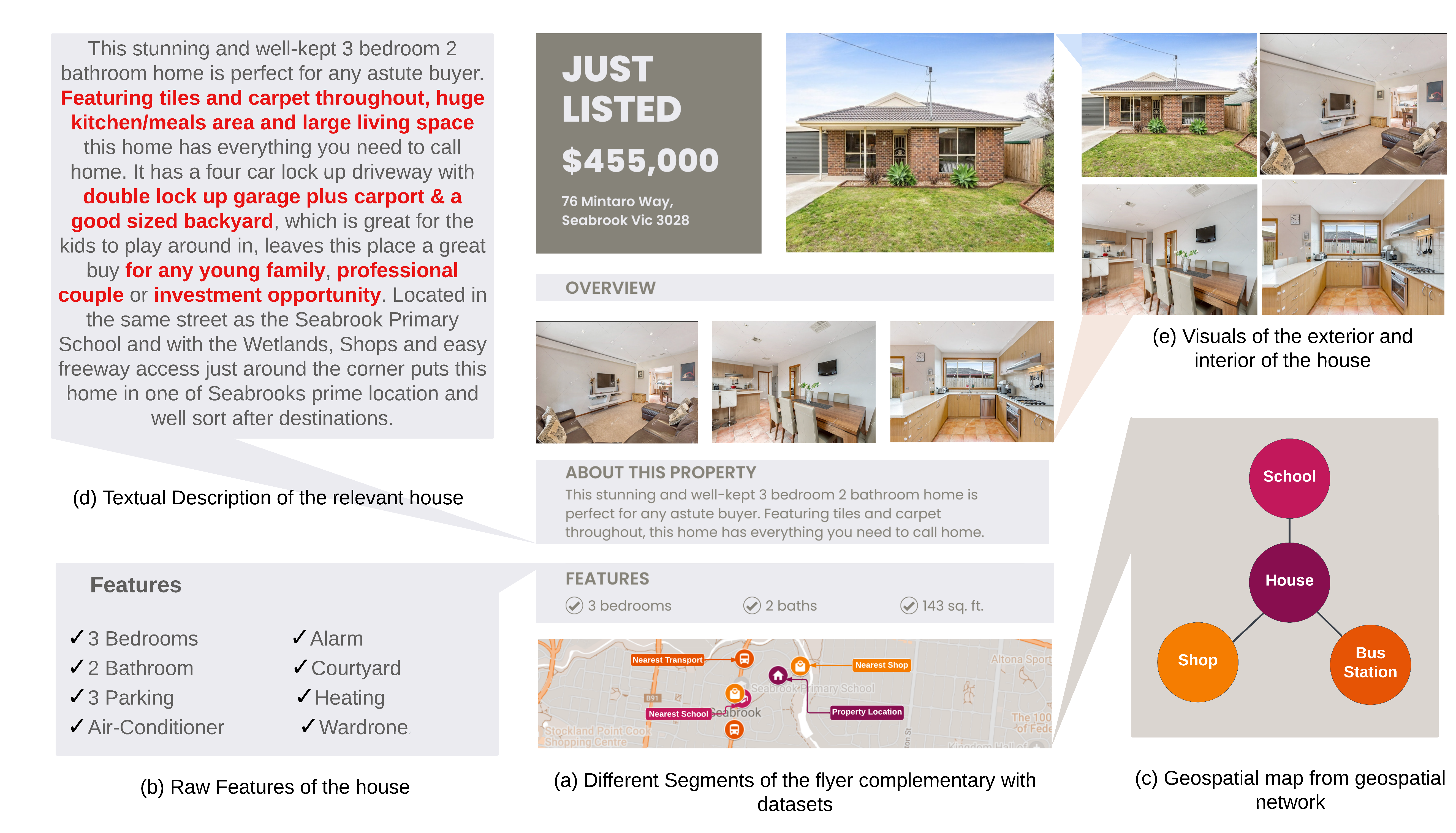}
        \caption{(a) A complete breakdown of a typical house flyer into different parts indicating our multi-dimensional dataset. (b) Different features of the house. (c) Geo-spatial network centered around the house where the edges are given with the nearby schools, shops and bus stations with their distance measured. (d) Textual description of the house in the advertisement describing the house. (e) The images of the house with both interior and exterior views. }
        \label{fig:data}
    %\end{subfigure}
\end{figure}

\subsection{Text Embedding}
\label{text embedding}

The detailed descriptions of houses, presented in plain text in an advertisement, offer an abundance of intricate information that are absent in the listed house raw features. Figure~\ref{fig:data}(d) shows some red-colored texts that cannot be captured by the raw features of the houses. These features that include some intricate details of some attribute, aesthetic features, etc.,  play a significant role in influencing the prices of various houses. To capture these extra features, we can employ a text embedding model to generate representations of the house description as one of the key features of our house prediction model.

In this context, we delve into the utilization of text embedding techniques. We first provide a brief introduction to the BERT language model, and subsequently, give a brief overview the state-of-the-art sentence embedding approach, SBERT, for the purpose of embedding our house descriptions.

BERT, as introduced by Devlin et al. \cite{devlin2018bert}, represents a multilayer bidirectional transformer network, building upon the foundational work outlined in Vaswani et al.'s \cite{vaswani2017attention}. BERT has consistently delivered state-of-the-art performance across a spectrum of natural language processing (NLP) tasks. For sentence-pair regression tasks, BERT's input typically comprises a sequence of two sentences. Each sequence begins with a specialized classification token ([CLS]) and is separated by a special token ([SEP]). The model incorporates a multi-head attention mechanism across its layers or transformer blocks. The output from this mechanism is then passed through a straightforward regression function to derive the final label.

\begin{figure}
        \centering
         \includegraphics[width=0.7\textwidth]{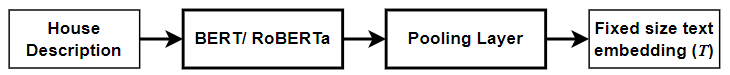}
         \caption{ SBERT architecture for deriving fixed length sentence embedding. }
         \label{fig:text2}
\end{figure}

SBERT~\cite{reimers2019sentence} builds upon the BERT architecture and incorporates advancements such as RoBERTa to enhance sentence embedding capabilities. We use SBERT to get the fixed size embedding vector of our house description (as depicted in Figure \ref{fig:text2}).

To obtain a fixed-size vector embedding, three commonly employed pooling strategies are typically employed. The first involves using the output of the CLS-token; the second computes the mean of all output vectors, referred to as the MEAN-strategy; the third approach calculates the maximum value over time among the output vectors, known as the MAX-strategy. Thus, given any house description $s$, the pooling layer gives a fixed size sentence embedding $T_E$.

\begin{equation}
\label{eqt2}
T_E = Pool (E_{enc}(s))
\end{equation}

\subsection{Image Embedding}
\label{image embedding}

In this section, we explain how the available images can be used as a potential house features for house price prediction. House images provide clear visual evidence of a property's condition, layout and surroundings. They also evoke emotion and aspiration in potential buyers or renters, making the decision-making process more informed. As an illustration, in Figure \ref{fig:data}(e), we can see that combining house images with textual descriptions provides a clearer understanding of the house's interior.

We use a multimodal language model known as Contrastive Learning Image Pretraining (CLIP) \cite{li2021supervision} to extract image embeddings from the pictures of houses. CLIP utilizes a joint training strategy to simultaneously train both image and text encoders. The objective of this integrated training is to guide the image encoder towards generating embeddings that closely align with the text embeddings. 

To train the image encoder $I_{enc}$, we employ the self-supervision technique~\cite{gomez2017self} within the CLIP framework, which involves training a batch of $b$ (image, text) pairs, with CLIP determining which of the $b\times b$ possible pairs correspond to each other. To achieve this, the text and image encoders are jointly trained to maximize the cosine similarity of embeddings for the $b$ real pairs in the batch, while minimizing the cosine similarity of embeddings for the $b^2-b$ incorrect pairs. To supervise the image encoder, we utilize a text encoder $T_{enc}$ based on the distilBert model~\cite{sanh2019distilbert}, while the image encoder is based on the Resnet50 model~\cite{he2016deep}. We optimize a symmetric cross-entropy loss function based on these similarity scores.

\begin{figure}
    \includegraphics[width=\textwidth]{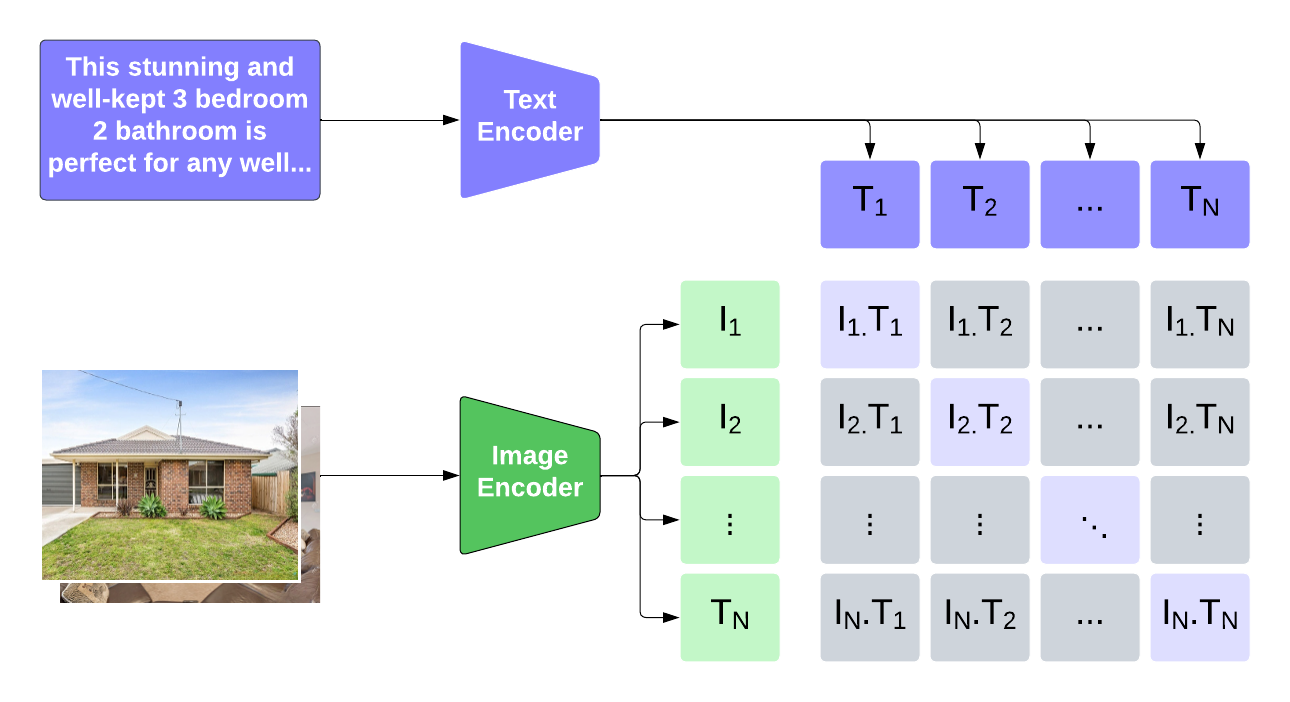}
         \caption{  CLIP \cite{li2021supervision} jointly trains an image encoder and a text encoder to predict the correct pairings of a batch of (image, text) training samples.}
         \label{fig:clip_training}
\end{figure}

The image encoder $I_{enc}$ generates an encoded representation $I_f$ for each batch of aligned images $I$, while the text encoder $T_{enc}$ creates an encoded representation $T_f$ for each batch of aligned textual descriptions $T$ of the houses.

\begin{equation}
\label{eq1}
I_f = I_{enc} (I)
\end{equation}

\begin{equation}
\label{eq2}
T_f = T_{enc} (T)
\end{equation}

After finding the encoded image feature $I_f$ and text feature $T_f$, the image projector $I_{proj}$ and the text projector $T_{proj}$ are used to embed them into the same dimension. This involves employing the image projector $I_{proj}$ to generate an image embedding $I_E$ from the encoded image feature $I_f$ and the text projector $T_{proj}$ to produce a text embedding $T_E$ from the encoded text feature $T_f$.

The optimization of a symmetric cross-entropy loss function based on similarity scores is performed later. This involves the computation of the cosine similarity between the embeddings of the image and text, as depicted by the following equation:

\begin{equation}
\label{eq5}
\text{sim}(i,t) = \frac{I_e \cdot T_e}{\left\lVert I_e \right\rVert \left\lVert T_e \right\rVert}
\end{equation}

where $I_e$ and $T_e$ are the embeddings for the image and text features, respectively.

Subsequently, we utilize the symmetric cross-entropy loss function to optimize the joint embedding. The loss function is computed according to the following equation:

\begin{equation}
\label{eq6}
\mathcal{L}(y, \hat{y}) = -\frac{1}{N}\sum_{i=1}^{N}\sum_{j=1}^{C}y_{i,j}\log(\hat{y}_{i,j})
\end{equation}

where $\mathcal{L}$ represents the loss function, $y$ is the true label or ground truth, $\hat{y}$ is the predicted label or output of the neural network, $N$ represents number of total samples in the dataset, $C$ denotes the total number of classes (i.e., classes representing different house price ranges) in the dataset, $i$ is the index of the sample being considered (ranging from 1 to $N$), $j$ is the index of the class being considered (ranging from 1 to $C$), $y_{i,j}$ denotes the true probability of the $i$-th sample belonging to class $j$ and $\hat{y}_{i,j}$ represents the predicted probability of the $i$-th sample belonging to class $j$.

\subsection{Fusing Diverse Embeddings and Raw Features}
\label{concatenation}
In the preceding subsections, we obtain geospatial network embedding ($G_E$), text embedding ($T_E$) and image embedding ($I_E$). Additionally,  we have the raw features of houses, identified as $F_{raw}$.

To have  a unified input in our downstream task of house price prediction, we concatenate all these embeddings. This technique involves the amalgamation of these distinct data sources by appending their numerical vectors in a specific order while preserving the original dimensionality of each embedding. As a result, a concatenated embedding vector, denoted as V, is generated, effectively consolidating information from various modalities.

Mathematically, the concatenated embedding $V$ is expressed as:

\begin{equation}
\label{eq7}
V = F_{\text{raw}} | G_E | T_E | I_E
\end{equation}

Subsequently, this  final embedding vector $V$ is used as the input for our house price prediction model to enhance the accuracy of our predictions.
\section{Experiments}
\label{experiments}
We present an extensive experimental evaluation of our proposed MHPP methodology using real estate dataset from Melbourne, Australia.

To assess the effectiveness of our method, we compare the performance of MHPP against  the state-of-the-art GSNE based approach\cite{das2021boosting}. Our selection of downstream regression models includes some of the top performers in Kaggle's house price prediction competitions \cite{one}, recent models for house price prediction \cite{chaturvedi2021real, xin2018modelling, xiong2019improve}, as well as renowned regression models like LightGBM \cite{ke2017lightgbm}, XGBoost \cite{chen2016xgboost} and Gradient Boosting \cite{chaturvedi2021real}.

%Our study's findings highlight MHPP's superiority over baseline models and its outperformance of state-of-the-art alternatives. 
Our approach achieves state-of-the-art performance, significantly reducing both mean absolute error and root mean squared error. An ablation study underscores the critical role of text and image embeddings in improving the accuracy of house price predictions.

\subsection{Dataset Description}
\label{subsec:data}
Our experimentation was conducted using a dataset comprising records of house transactions sourced from a prominent real estate website\footnote{https://www.realestate.com.au/}. The dataset encompasses real estate transactions in Melbourne, Australia's second-largest city in terms of population. It encompasses a total of 52,851 house transaction records from year 2013 to 2015. Furthermore, the dataset includes valuable details about nearby Points of Interest (POIs), encompassing regions, schools and train stations. This dataset comprehensively covers information related to 13,340 regions, 709 schools and 218 train stations. Additionally, each record within the dataset includes a concise textual description and images of the houses.

\subsubsection{House and POI Features}
Our dataset encompasses an extensive array of house attributes for each property, totaling 43 distinct features as outlined in Table \ref{tab:features}. These features span a spectrum from fundamental characteristics such as the number of bedrooms and bathrooms to intricate facility-related attributes like the presence of air-conditioning and tennis courts. Moreover, the dataset offers comprehensive insights into Melbourne's various regions at the SA1 level, incorporating details such as population count, average age, median personal income, and educational qualifications.

Furthermore, the dataset is enriched with information regarding different Points of Interest (POIs), including schools and train stations. These POIs are further categorized and accompanied by timetable data and precise location information. %This extensive POI information is invaluable for understanding house price determinants, prompting us to incorporate comprehensive features for each school and train station within the dataset.

\begin{table}[htbp]
\small
\centering
\begin{tabular}{|p{5cm}|p{5cm}|}
\hline
Number of bedrooms & Fireplace \\
Number of bathrooms & Fully fenced \\
Parking & Gas heating \\
Property type & Gym \\
Transaction date & Heating \\
Agency & Intercom \\
Latitude & Laundry \\
Longitude & Mountain \\
Air Conditioning & Park \\
Alarm & Swimming pool \\
Balcony & Renovated \\
BBQ & River view \\
City view & Rumpus room \\
Adjacency to schools & Sauna  \\
Adjacency to shops & Study rooms \\
Adjacency to transport & Sunroom \\
Courtyard & System heating \\
Number of dining rooms & Tennis court \\
Dish wash & Water views \\
Ducted & Wardrobe \\
Ensuite & Total additional features \\
Family room & \\
\hline
\end{tabular}
\caption{House Features}
\label{tab:features}
\end{table}

\subsubsection{House Descriptions}
The dataset also provides textual descriptions for each of the houses, capturing various aesthetics and features that may not be readily quantified. These descriptions exhibit varying lengths, with some extending up to a maximum of 280 words. An illustrative example of a textual description from the dataset is shown in Figure \ref{fig:data}(d).

\subsubsection{House Images}
Each property in the dataset typically features an average of five distinct images. These images collectively portray both the interior and exterior aspects of the houses, as exemplified in Figure \ref{fig:data}(e). It is  worth noting that while some houses may have been missing one image within this five-image set, we mitigated this by duplicating one of the four available images to maintain consistency. In our setting, we used the collage of five distinctive images to learn the correlation between house images and descriptions.

\subsection{Performance Metrics}
\label{subsec:metrics}
We adopt the mean absolute error (MAE) and root mean squared error (RMSE) as the evaluation metrics for assessing the performance of our model. MAE measures the average absolute difference between the predicted and ground truth values, providing insights into the overall accuracy of the model. On the other hand, RMSE gives more weight to larger errors by calculating the square root of the average squared differences. By utilizing these established metrics, we can effectively evaluate the efficacy of our proposed model in predicting house prices.

The formula for MAE is as follows:
\[
\text{{MAE}} = \frac{1}{N} \sum_{i=1}^{N} |z_i - \hat{z}_{ij}|
\]

Here, $N$ represents the number of samples, $z_i$ is the ground truth price and $\hat{z}_{ij}$ is the predicted house price.

The formula for RMSE is given by:
\[
\text{{RMSE}} = \sqrt{\frac{1}{N} \sum_{i=1}^{N} (z_i - \hat{z_i})^2}
\]

Here, $N$ represents the number of samples, $z_i$ is the ground truth price and $\hat{z}_{ij}$ is the predicted house price.

\subsection{Experimental Setup and Model Building}
\label{subsec:expsetup} 
In our research, a 12GB NVIDIA Tesla P100 GPU machine equipped with 16GB of memory and a Google Colab machine featuring 16 NVIDIA Tesla T4 GPUs were employed for model training. Dataset partitioning was conducted utilizing stratified random sampling, allocating 80\% for unsupervised training across three embedding models (namely, GSNE, text, and image), and reserving 20\% for subsequent model evaluation. Given that the test set remains unseen during training, favorable performance therein implies robust generalization of the embedding models to previously unobserved data, thereby highlighting the model's inductive capacity.
% We train three different models: GSNE for geo-spatial neighborhood, BERT for textual description and CLIP for house images to get the joint embedding of the house features. We use a 12GB NVIDIA Tesla P100 GPU machine with 16GB of memory and a 16 Nvidia Tesla T4 GPU Google Colab machine to train our model. In each case, we divide our dataset into two parts: training 80\% and test 20\% using stratified random sampling based approach.

\subsection{House Price Prediction Results}
\label{subsec:results}
We evaluate the effectiveness of our proposed MHPP approach, where we use house raw features, GSNE features, textual description and image features. As GSNE +raw features show the state-of-the-art of performance, we consider this as our baseline. Note that, GSNE uses raw features and apply first \& second order GNN embedding to form the final feature space to predict the house price.

To test the effectiveness of our MHPP  approach, we progressively add different text and image features in our house price prediction task. Note that, 
Specifically, we compare the performance of the regression models trained using (1) raw features, (2) raw + GSNE (first order), (3) raw + GSNE (second order), (4) raw + GSNE (first \& second order), (5) raw+ GSNE + text embedding, (6) raw + GSNE + image embedding and (7) raw + GSNE + text embedding + image embedding. 

We compare the performance using a wide range of regression models, including widely-used models such as Lasso and Ridge regression, Random Forest Regression, Elastic Net Regression,  Kernel-Ridge Regression, Gradient Boosting, XGBoost, and LightGBM.

\begin{table}[htbp]
    \centering
    \resizebox{\textwidth}{!}{
    \begin{tabular}{|l|l|l|l|l|l|l|l|}
    \hline
    \textbf{Metric} & \textbf{Method} & \textbf{Lasso} & \textbf{ENET} & \textbf{KRR} & \textbf{GBOOST} & \textbf{XGB} & \textbf{LGBM} \\ 
    \hline
    & (1) Raw & 0.251 & 0.25 & 0.149 & 0.136 & 0.143 & 0.135 \\ 
    & (2) Raw+First & 0.22 & 0.216 & 0.141 & 0.128 & 0.133 & 0.129 \\ 
    & (3) Raw+Second & 0.198 & 0.196 & 0.14 & 0.131 & 0.136 & 0.131 \\ 
    MAE& (4) Raw+first+second & 0.209 & 0.205 & 0.135 & 0.125 & 0.132 & 0.127 \\ 
    & (5) Raw+first+Second+text & 0.175 & 0.166 & 0.123 & 0.12 & 0.124 & 0.119 \\ 
    & (6) Raw+first+Second+image & 0.165 & 0.161 & 0.127 & 0.119 & 0.12 & 0.117 \\ 
    & \textbf{(7) Raw+first+Second+text+image} & \textbf{0.159} & \textbf{0.151} & \textbf{0.12} & \textbf{0.116} & \textbf{0.115} & \textbf{0.112} \\
    \hline
    & (1) Raw & 0.333 & 0.331 & 0.206 & 0.195 & 0.2 & 0.19 \\ 
    & (2) Raw+First & 0.295 & 0.291 & 0.196 & 0.184 & 0.188 & 0.182 \\ 
    & (3) Raw+Second & 0.277 & 0.271 & 0.194 & 0.188 & 0.191 & 0.184 \\ 
    RMSE & (4) Raw+first+second & 0.29 & 0.289 & 0.19 & 0.181 & 0.187 & 0.18 \\ 
    & (5) Raw+first+Second+text & 0.244 & 0.23 & 0.171 & 0.173 & 0.174 & 0.168 \\ 
    & (6) Raw+first+Second+image & 0.233 & 0.224 & 0.178 & 0.173 & 0.172 & 0.167 \\ 
    & \textbf{(7) Raw+first+Second+text+image} & \textbf{0.223} & \textbf{0.211} & \textbf{0.168} & \textbf{0.169} & \textbf{0.164} & \textbf{0.16} \\ 
    \hline
    \end{tabular}
    }
    \caption{Performance Comparison of House Price Prediction}
    \label{tab:result1}
\end{table}

\begin{table}[htbp]
\centering
 \resizebox{\textwidth}{!}{
\begin{tabular}{|l|l|l|l|l|l|l|l|}
\hline

& &\multicolumn{6}{|c|}{Percentage of Improvement(\%)}\\
\hline

Metric &  Method & Lasso & ENET & KRR & GBOOST & XGB & LGBM \\ \hline

& Raw + first +second &  \multicolumn{6}{|c|}{Baseline}\\ 
\hhline{|~|~|------|}
& Raw+first+Second+text & 16.27 & 19.02 & 8.89 & 4 & 6.06 & 6.3 \\ 
MAE& Raw+first+Second+image & 21.05 & 21.46 & 5.93 & 4.8 & 9.09 & 7.87 \\ 
& Raw+first + Second + text+image & 23.92 & 26.34 & 11.11 & 7.2 & 12.88 & 11.81 \\ 
 \hline

& {Raw + first +second} &\multicolumn{6}{|c|}{Baseline}\\
\hhline{|~|~|------|}
&Raw+first+Second+text & 15.86 & 20.42 & 10 & 4.42 & 6.95 & 6.67 \\ 
RMSE & Raw+first+Second+image & 19.66 & 22.49 & 6.32 & 4.42 & 8.02 & 7.22 \\ 
&Raw+first + Second + text+image & 23.1 & 26.99 & 11.58 & 6.63 & 12.3 & 11.11 \\ 
\hline
\end{tabular}
}
\caption{Percentage(\%) of improvement for each method}
\label{tab:result2}
\end{table}

\begin{figure}
  \centering
  \begin{tabular}{@{}c@{}}
    \includegraphics[width=.98\linewidth,height=150pt]{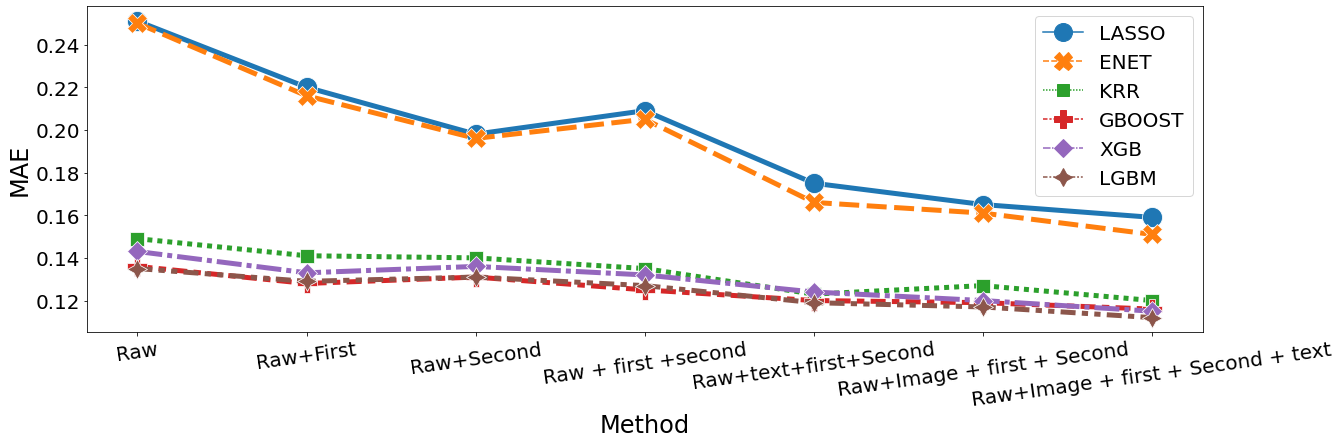} \\[\abovecaptionskip]
    \small (a)
  \end{tabular}

  \vspace{\floatsep}

  \begin{tabular}{@{}c@{}}
    \includegraphics[width=.98\linewidth,height=150pt]{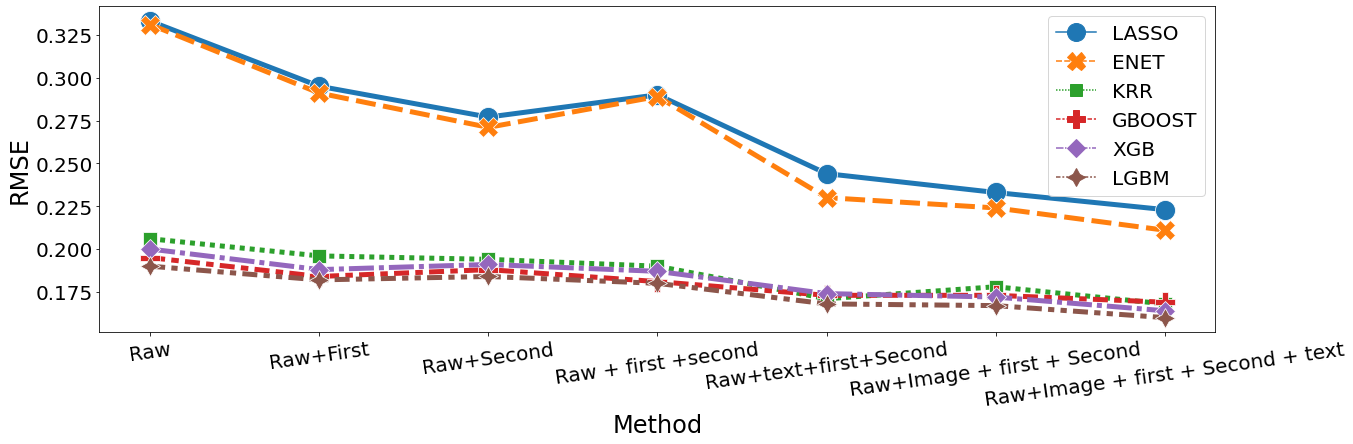} \\[\abovecaptionskip]
    \small (b)
  \end{tabular}

  \caption{(a)Mean Absolute Error(MAE) for different models and methods,
     (b)Root Mean Square Error(RMSE) for different models and methods}\label{fig:res}
\end{figure}

\subsubsection{Result Summary}
Table \ref{tab:result1} compares the performance of different prediction models, including Lasso, ENET, KRR, GBOOST, XGB and LGBM, using metrics such as MAE (Mean Absolute Error) and RMSE (Root Mean Square Error). The table showcases the performance of each model under different feature combinations, ranging from raw data to the inclusion of text and image features. The best-performing method for each metric is bolded.

The results reveal that our MHPP method, i.e., ``Raw+first+Second+text+image", consistently outperforms other models, achieving the lowest MAE (0.159) and RMSE (0.223) values. It demonstrates the importance of considering multiple features, including both textual and visual information, for accurate housing price estimation.

In Figure~\ref{fig:res}(a) and Figure~\ref{fig:res}(b), we showcase MAE and RMSE for a variety of regression models trained using diverse feature embedding combinations. Our analysis underscores that regression models trained with a comprehensive feature embedding set, including raw data, images, first-order embedding, second-order embedding, and text embedding, consistently attain the lowest error rates. The horizontal axis in the figure denotes the distinct feature combinations employed during model training, while the vertical axis represents the associated error values.

Table \ref{tab:result2} focuses on quantifying the improvement achieved by each method compared to the baseline method,``Raw+first+second." The improvement percentages are calculated for both MAE and RMSE metrics, indicating how each model enhances the prediction accuracy.

The ``Raw+first+second+text+image" method stands out once again, exhibiting the highest improvement percentages across all models. It showcases a 23.92\% improvement in MAE and a 23.1\% improvement in RMSE compared to the baseline, respectively. This emphasizes the significance of integrating text and image features, as it leads to substantial enhancements in housing price prediction accuracy.

\subsubsection{Impact of Text Embedding}
The impact of text embedding on housing price prediction is evident when considering the data provided. The results clearly demonstrate that text embedding plays a crucial role in improving the accuracy and performance of the prediction task. Without text embedding, the models show suboptimal performance, as observed in Table \ref{tab:result1}. However, when text embedding is incorporated into the models, there is a significant improvement in predictive capability, as shown in Table \ref{tab:text-embedding}. 

Analyzing the results from \ref{tab:result2}, we observe that the inclusion of text embedding, along with various combinations of raw data, GSNE (1st order / 2nd order / both) and image data, leads to substantial enhancements in the predictive accuracy. The MAE and RMSE values consistently decrease after the addition of text embedding, indicating a higher level of precision in the housing price predictions. 

\begin{table}[htbp]
    \centering
    \resizebox{\textwidth}{!}{
    \begin{tabular}{|l|l|l|l|l|l|l|l|}
    \hline
    \textbf{Metric} & \textbf{Method} & \textbf{Lasso} & \textbf{ENET} & \textbf{KRR} & \textbf{GBOOST} & \textbf{XGB} & \textbf{LGBM} \\ 
    \hline
     & Raw+Text & 0.216 & 0.207 & 0.132 & 0.137 & 0.13 & 0.125 \\ 
     & Raw+First+Text & 0.195 & 0.184 & 0.126 & 0.122 & 0.124 & 0.12 \\ 
    MAE & Raw+Second+Text & 0.181 & 0.172 & 0.125 & 0.125 & 0.126 & 0.122 \\ 
     & Raw+first+Second+Text & 0.175 & 0.166 & 0.123 & 0.12 & 0.124 & 0.119 \\ 
     & Raw+first+Second+image+Text & 0.159 & 0.151 & 0.12 & 0.116 & 0.115 & 0.112 \\
     \hhline{|~|-------|}
     & Percentage of Average Improvement(\%) & +10.76 & +13.9 & +9.43 & +3.01 & +6.69 & +6.37 \\
    \hline
     & Raw+Text & 0.29 & 0.278 & 0.183 & 0.195 & 0.182 & 0.176 \\ 
     & Raw+First+Text & 0.263 & 0.25 & 0.175 & 0.176 & 0.175 & 0.17 \\ 
    RMSE & Raw+Second+Text & 0.255 & 0.24 & 0.174 & 0.179 & 0.178 & 0.171 \\ 
     & Raw+first+Second+Text & 0.244 & 0.23 & 0.171 & 0.173 & 0.174 & 0.168 \\ 
     & Raw+first+Second+image+Text & 0.223 & 0.211 & 0.168 & 0.169 & 0.164 & 0.16 \\
     \hhline{|~|-------|}
     & Percentage of Average Improvement(\%) & +10.37 & +13.55 & +9.56 & +3.17 & +6.86 & +6.38 \\
    \hline
    \end{tabular}
    }
    \caption{Effect of Text Embedding on Housing Price Prediction Performance.}
    \label{tab:text-embedding}
\end{table}

\subsubsection{Impact of Image Embedding}
The incorporation of image embedding brings about a significant improvement in the housing price prediction task. The data provided clearly demonstrates the impact of image embedding on enhancing the performance and accuracy of the prediction models. When image embedding is not utilized, the predictive capability of the models is observed to be limited, as evident from the results presented in Table \ref{tab:result1}. However, upon integrating image embedding into the models, there is a drastic improvement in their performance, as highlighted in Table \ref{tab:image-embedding}.

\begin{table}[htbp]
    \centering
    \resizebox{\textwidth}{!}{%
    \begin{tabular}{|l|l|l|l|l|l|l|l|}
    \hline
    \textbf{Metric} & \textbf{Method} & \textbf{Lasso} & \textbf{ENET} & \textbf{KRR} & \textbf{GBOOST} & \textbf{XGB} & \textbf{LGBM} \\ 
    \hline
    & Raw+Image & 0.204 & 0.201 & 0.137 & 0.143 & 0.129 & 0.125 \\ 
    & Raw+First+Image & 0.189 & 0.183 & 0.13 & 0.122 & 0.121 & 0.117 \\ 
    & Raw+Second+Image & 0.166 & 0.162 & 0.129 & 0.12 & 0.121 & 0.118 \\ 
    MAE& Raw+Text+Image & 0.193 & 0.186 & 0.128 & 0.14 & 0.122 & 0.119 \\ 
    & Raw+First+Second+Image & 0.165 & 0.161 & 0.127 & 0.119 & 0.12 & 0.117 \\ 
    & Raw+First+Second+Text+Image & 0.159 & 0.151 & 0.12 & 0.116 & 0.115 & 0.112 \\ 
    \hhline{|~|-------|}
     & Percentage of Average Improvement(\%) & +14.97 & +15.48 & +5.85 & +2.31 & +8.72 & +7.53 \\
    \hline
    & Raw+Image & 0.276 & 0.27 & 0.191 & 0.202 & 0.183 & 0.177 \\ 
    & Raw+First+Image & 0.257 & 0.249 & 0.182 & 0.177 & 0.172 & 0.168 \\ 
    & Raw+Second+Image & 0.234 & 0.225 & 0.181 & 0.175 & 0.173 & 0.168 \\ 
    RMSE & Raw+Text+Image & 0.262 & 0.252 & 0.179 & 0.197 & 0.173 & 0.169 \\ 
    & Raw+First+Second+Image & 0.233 & 0.224 & 0.178 & 0.173 & 0.172 & 0.167 \\ 
    & Raw+First+Second+Text+Image & 0.223 & 0.211 & 0.168 & 0.169 & 0.164 & 0.16 \\ 
    \hhline{|~|-------|}
     & Percentage of Average Improvement(\%) & +13.91 & +14.99 & +5.23 & +2.14 & +7.52 & +6.53 \\
    \hline
    \end{tabular}%
    }
    \caption{Effect of image embedding on performance metrics}
    \label{tab:image-embedding}
\end{table}

\subsubsection{Effect of Embedding Dimension}
%\label{subsec:embedding}
Choosing the appropriate embedding dimension is an important task. It is essential to strike a balance between the embedding size and the model's runtime and memory requirements while maintaining the house price prediction performance. We evaluate the impact of various embedding dimensions on model performance and select the one that provides the best trade-off between accuracy and efficiency. 

\subsubsection{Selection of Text Embedding Dimension}
We utilized Principal Component Analysis (PCA), a dimensionality reduction technique \cite{ivosev2008dimensionality} to compute text embedding of various dimensions from default length of 768. Our experiment shows that downsampling too much can cause information loss, hence the apparent increase in MAE and RMSE as shown in Table \ref{tab:embedding-performance-text}. Thus, we determined that an embedding size of 128 gives the most promising result, while maintaining performance comparable to the original dimension.

\begin{table}[htbp]
\centering
\begin{tabular}{@{}|c|c|c|@{}}
\hline
Embedding Dimension & MAE & RMSE \\
\hline
32 & 0.218 & 0.293 \\
64 & 0.216 & 0.290 \\
\textbf{128} & \textbf{0.215} & \textbf{0.289} \\
768  & 0.220 & 0.296 \\
\hline
\end{tabular}
\caption{Effect of Dimension on text embedding}
\label{tab:embedding-performance-text}
\end{table}

\subsubsection{Selection of Image Embedding Dimension}
Through experimentation with different embedding dimensions for image data, it was found that utilizing an embedding dimension of 256 produced the highest level of performance for the model. The results presented in Table \ref{tab:embedding-performance-img} further support this finding, as they demonstrate that the 256-dimensional embedding outperformed other dimensions in terms of accuracy and other evaluation metrics. 

\begin{table}[htbp]
\centering
\begin{tabular}{@{}|c|c|c|@{}}
\hline
Embedding Dimension & MAE & RMSE \\
\hline
32  & 0.250 & 0.331 \\
64 & 0.247 & 0.327 \\
128 & 0.247 & 0.328 \\
\textbf{256} & \textbf{0.204} & \textbf{0.276} \\
512 & 0.207 & 0.279 \\
\hline
\end{tabular}
\caption{Effect of Dimension on image embedding}
\label{tab:embedding-performance-img}
\end{table}
\section{Conclusion}
\label{conclusion}
In this paper, we have proposed a MHPP model that incorporates a wide range of data types, which include house raw features, geo-spatial neighborhood, house description and house pictures, for an enhanced highly accurate house price prediction system. In particular, our model outperformed the state-of-the-art house price prediction methods that work on house attributes and geo-spatial neighborhood as their features. We have exploited the the power of transformer based language model, SBERT and multi-modal language model CLIP to extract features from house description and images, respectively. Then, we have used these learned features, along with house raw-features and geo-spatial neighborhood features, to have a unique embedding to represent a house.  Our extensive experimentation has demonstrated that our MHPP model can improve the accuracy of the house price prediction significantly, irrespective of the choice of downstream regression model and by a maximum of 26.34\%(MAE) and 26.99\%(RMSE) compared to the base state-of-the-art model. Our study can further help improve important real estate assessments which include purchase suggestions and choosing appropriate alternatives suitable for the buyer needs. In future, the impact of many complex correlations of various real life data including social security and trending assets of social media can further be exploited to better capture the essence of house price prediction in a multi-modal architecture.

 \section*{Declarations}

% Some journals require declarations to be submitted in a standardised format. Please check the Instructions for Authors of the journal to which you are submitting to see if you need to complete this section. If yes, your manuscript must contain the following sections under the heading `Declarations':
\textbf{Funding: }

The authors did not receive support from any organization for the submitted work.

No funding was received to assist with the preparation of this manuscript.

No funding was received for conducting this study.

No funds, grants, or other support was received.

\noindent\textbf{Conflict of interest/Competing interests: }

The authors have no relevant financial or non-financial interests to disclose.

The authors have no conflicts of interest to declare that are relevant to the content of this article.

All authors certify that they have no affiliations with or involvement in any organization or entity with any financial interest or non-financial interest in the subject matter or materials discussed in this manuscript.

The authors have no financial or proprietary interests in any material discussed in this article.

\bibliography{0-main}% common bib file

\end{document}